\def\year{2019}\relax
\documentclass[letterpaper]{article} 
\usepackage{aaai19}  
\usepackage{times}  
\usepackage{helvet}  
\usepackage{courier}  
\usepackage{url}  
\usepackage{graphicx}  
\usepackage{amsmath}
\usepackage{amsfonts,amssymb}
\usepackage{xspace}
\usepackage{setspace}
\usepackage{stmaryrd}
\usepackage{comment}
\usepackage[english]{babel}
\usepackage[utf8]{inputenc}

\usepackage{adjustbox}
\usepackage[shortend,ruled,linesnumbered]{algorithm2e}
\usepackage{array}
\usepackage{booktabs}
\usepackage{multirow}
\usepackage{subcaption} 

\usepackage{enumitem}

\graphicspath{{figures/}}
\DeclareGraphicsExtensions{.pdf}

\newcommand{\fml}[1]{\ensuremath\mathcal{#1}}

\newcommand{\tn}[1]{\textnormal{#1}}
\newcommand{\tl}[1]{\textsl{#1}}

\DeclareMathOperator*{\entails}{\vDash}
\DeclareMathOperator*{\nentails}{\nvDash}

\newcommand{\stwop}{\Sigma_2^\tn{P}}

\newcommand{\papsd}{$P=(\fml{V},\fml{H},\fml{E},\fml{F})$\xspace}
\newcommand{\papdef}{$P=(\fml{V},\fml{H},\fml{E},\fml{F},c)$\xspace}

\newtheorem{definition}{Definition}

\newtheorem{example}{Example}

\newcommand{\mailtodomain}[1]{\href{mailto:#1@ciencias.ulisboa.pt}{\textrm{\nolinkurl{#1}}}}

\nocopyright

\def\TheTitle{Abduction-Based Explanations for Machine Learning Models}
\begin{document}
%

\title{\TheTitle
\thanks{
  This work was supported by FCT grants ABSOLV
  (LIS\-BOA-01-0145-FEDER-028986), FaultLocker
  (PTDC/CCI‐COM/29300/2017), SAFETY (SFRH/BPD/120315/2016), and
  SAMPLE (CEECIND/04549/2017).}
}

\author{
  Alexey Ignatiev$^{1,3}$, Nina Narodytska$^2$, Joao Marques-Silva$^1$ \\
  $^1$ Faculty of Science, University of Lisbon, Portugal \\
  $^2$ VMware Research, CA, USA \\
  $^3$ ISDCT SB RAS, Irkutsk, Russia \\
  \{aignatiev,jpms\}@ciencias.ulisboa.pt, nnarodytska@vmware.com
}
\maketitle

%
\begin{abstract} 
  The growing range of applications of Machine Learning (ML) in
  a multitude of settings motivates the ability of computing small 
  explanations for predictions made. Small explanations are generally
  accepted as easier for human decision makers to understand.
  Most earlier work on computing explanations is based on heuristic
  approaches, providing no guarantees of quality, in terms of how
  close such solutions are from cardinality- or subset-minimal
  explanations.
  This paper develops a constraint-agnostic solution for computing
  explanations for any ML model. The proposed solution exploits
  abductive reasoning, and imposes the requirement that the ML model
  can be represented as sets of constraints using some target
  constraint reasoning system for which the decision problem can be
  answered with some oracle.
  The experimental results, obtained on well-known datasets, validate
  the scalability of the proposed approach as well as the quality of
  the computed solutions.
\end{abstract}
%

%

\section{Introduction} 

The fast growth of machine learning (ML) applications has motivated
efforts to validate the results of ML
models~\cite{narodytska-corr18,kwiatkowska-ijcai18,narodytska-ijcai18d,narodytska-aaai18,kwiatkowska-tacas18,kwiatkowska-cav17,barrett-cav17},
but also efforts to explain predictions made by such
models~\cite{grefenstette-jair18,rudin-aaai18,guestrin-aaai18,ipnms-ijcar18,muller-dsp18,guestrin-kdd16,leskovec-kdd16,muller-jmlr10}.
One concern is the application of ML models, including Deep Neural
Networks (DNNs) in safety-critical applications, and the need to
provide some sort of certification about correctness of operation.
The importance of computing explanations is further underscored by a
number of recent
works~\cite{goodman-aimag17,kim-corr17,monroe-cacm18,darwiche-cacm18,lipton-cacm18},
by recent regulations~\cite{eu-reg16}, ongoing research
programs~\cite{darpa-xai16}, but also by recent
meetings~\cite{ijcaiw17,icmlw17,nipsw17,floc18-smlmfm}.

For logic-based models, e.g.\ decision trees and sets, explanations
can be obtained directly from the model, and related research work has
mostly focused on minimizing the size of
representations~\cite{leskovec-kdd16,ipnms-ijcar18}.
Nevertheless, important ML models, that include neural networks (NNs),
support vector machines (SVMs), bayesian network classifiers (BNCs),
among others, 
do not naturally provide explanations to predictions made. Most work
on computing explanations for such models is based on heuristic
approaches, with no guarantees of
quality~\cite{guestrin-kdd16,leskovec-kdd16,rudin-kdd17,xing-corr18,rudin-aaai18}.
Recent work~\cite{xing-corr18} suggests that these heuristic
approaches can perform poorly in practical settings.
For BNCs, a recent
compilation-based approach \cite{darwiche-ijcai18}
represents a first step towards computing explanations with guarantees
of quality. Nevertheless, a drawback of compilation approaches is the
exponential worst-case size of the compiled representation, and also
the fact that it is specific to BNCs.

This paper explores a different path, and proposes a principled
approach for computing minimum explanations of ML models.
Concretely, the paper exploits abductive reasoning for computing
explanations of ML models with formal guarantees,
e.g.~cardinality-minimal or subset-minimal explanations. More
importantly, our approach exploits the best properties of logic-based
and heuristic-based approaches. Similar to heuristic approaches, our
method is model-agnostic. If an ML model can be expressed in a
suitable formalism then it can be explained in our framework. Similar
to logic-based approaches, our method provides formal guarantees on
the generated explanations. For example, we can generate
cardinality-minimal explanations. Moreover, it  allows a user to
specify custom constraints on explanations, e.g.\ a user might have
preferences over explanations.
%
%
%
%
Although the use of abductive reasoning for computing explanations is
well-known~\cite{shanahan-ijcai89,marquis-fair91}, its application in
explainable AI is novel to the best of our knowledge.
%
The abductive reasoning solution is based on representing the ML model
as a set of constraints in some theory (e.g.\ a decidable theory of
first-order logic).
The ML model prediction explanation approach proposed in this paper
imposes mild requirements on the target ML model and the constraint
reasoning system used. One must be able to encode the ML model as a
set of constraints, and the constraint reasoning system must be able to
answer entailment queries.

To illustrate the application of abductive reasoning for computing
explanations, the paper focuses on Neural Networks. As a result, a
recently proposed encoding of NNs into Mixed Integer Linear
Programming is used~\cite{fischetti-cj18}. This encoding is also
evaluated with Satisfiability Modulo Theories (SMT) solvers.
%
Although other recently proposed MILP encodings could be
considered~\cite{tedrake-corr17}, the most significant differences
are in the algorithm used.

The experimental results, obtained on representative problem
instances, demonstrate the scalability of the proposed approach, and
confirm that small explanations can be computed in practice.



\section{Background} 

\paragraph{Propositional Formulas, Implicants and Abduction.}
We assume definitions standard in propositional logic and
satifiability~\cite{sat-handbook09}, with the usual definitions for
(CNF) formulas, clauses and literals. Where required, formulas are
viewed as sets of clauses, and clauses as sets of literals.
For propositional formulas, a(n) (partial) assignment is a (partial)
map from variables to $\mathbb{B}=\{0,1\}$.
A satisfying assignment is such that the valuation of the formula
(under the usual semantics of propositional logic) is 1. Throughout
the paper, assignments will be represented as conjunctions of
literals.
Moreover,
let $\fml{F}$ be a propositional formula defined on a set of variables
$X=\{x_1,\ldots,x_n\}$. A literal is either a variable $x_i$ or its
complement $\neg x_i$. A term is a set of literals, interpreted as a
conjunction of literals. A term $\pi$ is an \emph{implicant} if
$\pi\entails\fml{F}$. An implicant $\pi$ is a \emph{prime implicant}
if $\pi\entails\fml{F}$, and for any proper subset
$\pi'\subsetneq\pi$, $\pi'\nentails\fml{F}$~\cite{marquis-hdrums00}.
A prime implicant $\pi$ given a satisfying assignment $\mu$ is any
prime implicant $\pi\subseteq\mu$.
Given a CNF formula and a satisfying assignment, a prime implicant can
be computed in polynomial time. For an arbitrary propositional
formula, given a satisfying assignment, a prime implicant can be
computed with a linear number of calls to an NP
solver (e.g.~\cite{liberatore-aij05}). 
In contrast, computing the shortest prime implicant is hard for
$\stwop$~\cite{liberatore-aij05}, the second level of the polynomial
hierarchy. 

Let $\fml{F}$ denote a propositional theory which, for the goals of
this paper can be understood as a set of clauses.
Let $\fml{H}$ and $\fml{E}$ be respectively a set of hypotheses and a
set of manifestations (or the evidence), which often correspond to
unit clauses, but can also be arbitrary clauses.

A propositional abduction problem (PAP) is a 5-tuple \papdef.
$\fml{V}$ is a finite set of variables. $\fml{H}$, $\fml{E}$ and
$\fml{F}$ are CNF formulas representing, respectively, the set of
hypotheses, the set of manifestations, and the background theory.
$c$ is a cost function associating a cost with each clause of $\fml{H}$,
$c:\fml{H}\to\mathbb{R}^{+}$.

Given a background theory $\fml{F}$, a set $\fml{S}\subseteq\fml{H}$
of hypotheses is an explanation (for the manifestations) if: (i)
$\fml{S}$ entails the manifestations $\fml{E}$ (given $\fml{F}$); and
(ii) $\fml{S}$ is consistent (given $\fml{F}$).
%
The propositional abduction problem consists in computing a minimum size
explanation for the manifestations subject to the background theory,
e.g.\ \cite{eiter-jacm95,jarvisalo-kr16,imms-ecai16}.

\begin{definition}[Minimum-size explanations for $P$]\label{def:expl}
Let \papdef be a PAP. The set of explanations of $P$ is given by the
set $\tn{Expl}(P) =\{\fml{S}\subseteq\fml{H}\:|\:\fml{F}\land\fml{S}\nentails\bot,\fml{F}\land\fml{S}\entails\fml{E}\}$.
The minimum-cost solutions of $P$ are given by $\tn{Expl}_c(P) =
\tn{argmin}_{E\in\tn{Expl}(P)}(c(E))$.
\end{definition}
Subset-minimal explanations can be defined similarly.
%
Moreover, throughout the paper the cost function assigns unit cost to
each hypothesis, and so we use the following alternative notation for a
PAP $P$, \papsd.

\paragraph{First Order Logic and Prime Implicants.}
%
We assume definitions standard in first-order logic
(FOL)~(e.g.~\cite{gallier-bk03}).
Given a signature $\fml{S}$ of predicate and function symbols, each of
which characterized by its arity, a theory $\fml{T}$ is a set of
first-order sentences over $\fml{S}$.
$\fml{S}$ is extended with the predicate symbol $=$, denoting logical
equivalence.
A model $\fml{M}$ is a pair $\fml{M}=(\fml{U},\fml{I})$, where
$\fml{U}$ denotes a universe, and $\fml{I}$ is an interpretation that
assigns a semantics to the predicate and function symbols of
$\fml{S}$.
A set $\fml{V}$ of variables is assumed, which is distinct from
$\fml{S}$.
A (partial) assignment $\nu$ is a (partial) function from $\fml{V}$
to $\fml{U}$. Assignments will be represented as conjunctions of
literals (or \emph{cubes}), where each literal is of the form $v=u$,
with $v\in\fml{V}$ and $u\in\fml{U}$. Throughout the paper, cubes and
assignments will be used interchangeably.
The set of free variables in some formula $\fml{F}$ is denoted by
$\tl{free}(\fml{F})$.
Assuming the standard semantics of FOL, and given an assignment $\nu$
and corresponding cube $C$, the notation $\fml{M},C\entails\fml{F}$ is
used to denote that $\fml{F}$ is true under model $\fml{M}$ and cube
$C$ (or assignment $\nu$). In this case $\nu$ (resp.~$C$) is referred
to as satisfying assignment (resp.~cube), with the assignment being
partial if $|C|<|\fml{V}|$ (and so if $\nu$ is partial).
A solver for some FOL theory $\fml{T}$ is referred to as a
$\fml{T}$-oracle.

A well-known  generalization of prime implicants to
FOL~\cite{marquis-fair91} will be used throughout.

\begin{definition}
  Given a FOL formula $\fml{F}$ with a model
  $\fml{M}=(\fml{U},\fml{I})$, a cube $C$ is a prime implicant of
  $\fml{F}$ if
  \begin{enumerate}
  \item $\fml{M},C\entails\fml{F}$.
  \item If $C'$ is a cube such that $\fml{M},C'\entails\fml{F}$ and
    $\fml{M},C'\entails C$, then $\fml{M},C\entails C'$.
  \end{enumerate}
\end{definition}
A smallest prime implicant is a prime implicant of minimum size.
Smallest prime implicants can be related with minimum satisfying
assignments~\cite{dillig-cav12}.
Finally, a prime implicant $C$ of $\fml{F}$ and $\fml{M}$ given a cube
$C'$ is a prime implicant of $\fml{F}$ such that $C\subseteq C'$.

Satisfiability Modulo Theories (SMT) represent restricted (and often
decidable) fragments of FOL~\cite{sebastiani-hdbk09a}. All the    
definitions above apply to SMT.

\paragraph{Mixed Integer Linear Programming~(MILP).}
In this paper, a MILP is defined over a set of variables $V$, which
are partitioned into real (e.g.\ $Y$), integer (e.g.\ $W$) and Boolean
(e.g.\ $Z$) variables.
%
\begin{equation}
  \begin{array}{lll}
    \tn{min} & \sum_{v_j\in V}c_jv_j\\[5pt]
    \tn{s.t.} & \sum_{v_j\in V}a_{ij}v_j\le b_i & 1\le i\le K\\
  \end{array}
\end{equation}
where $\{b_1,\ldots,b_K\}$ can either be real, integer or Boolean
values.
To help with the encoding of ML models, we will exploit indicator
constraints
(e.g.~\cite{fischetti-coa16,fischetti-cj18}) of the form:
\begin{equation}
l_i \to \left(\sum_{v_j\in V}a_{ij}v_j\le b_i\right)
\end{equation}
where $l_i$ is some propositional literal.

Clearly, under a suitable definition of signature $\fml{S}$ and model
$\fml{M}$, with $\fml{U}\triangleq\mathbb{R}\cup\mathbb{Z}\cup\mathbb{B}$,
(smallest) prime implicants can be computed for MILP.

\paragraph{Minimal Hitting Sets.}
Given a collection $\mathbb{S}$ of sets from a universe $\mathbb{U}$,
a hitting set $h$ for $\mathbb{S}$ is a set such that
$\forall S \in \mathbb{S}, h \cap S \ne \emptyset.$
A hitting set $h$ is said to be \emph{minimal} if none of its subsets
is a hitting set.



\section{ML Explanations as Abductive Reasoning} 

We consider the representation of an ML model using a set of
constraints, represented in the language of some constraint reasoning
system. Associated with this constraint reasoning system, we assume
access to an oracle that can answer entailment queries. For example,
one can consider Satisfiability Modulo Theories, Constraint
Programming, or Mixed Integer Linear Programming.

Moreover, we associate the quality of an explanation with the number
of specified features associated with a prediction. As a result, one
of the main goals is to compute cardinality-minimal
explanations. Another, in practice more relevant due to performance
challenges of the first goal, is to compute subset-minimal
explanations.

\paragraph{Propositional Case.}
Let $\mathbb{M}$ denote some ML model.
We assume a set of (binarized) features $\fml{V}=\{f_1,\ldots,f_k\}$,
and a classification problem with two classes $\{c_0,c_1\}$.
Let some $p\in\{c_0,c_1\}$ denote some prediction.
Moreover, let us assume that we can associate a logic theory $\fml{T}$
with the ML model $\mathbb{M}$, and encode $\mathbb{M}$ as a formula
$\fml{F}$.

Given the above, one can compute cardinality minimal explanations for
$p$ as follows.
Let $\fml{H}=\{(f_i),(\neg f_i)|f_i\in\fml{V}\}$, let
$\fml{E}=\{(p)\}$, and associate a unit cost function $\nu$ with each
unit clause of
$\fml{H}$.
Then, any explanation for the PAP
$P=(\fml{V},\fml{H},\fml{E},\fml{F})$ is an explanation for the
prediction $p$.
To compute cardinality minimal explanations, we can use for example
a recently proposed approach~\cite{imms-ecai16}.
Moreover, observe that if features are real-valued, then the approach
outlined above for computing cardinality minimal explanations does not
apply.

In a more concrete setting of some point $\phi$ in feature space and
prediction $p$, we consider a concrete set $\fml{H}$, where the value
of each feature of $\phi$ is represented with a unit clause.
As above, we can consider propositional abduction, which in this case
corresponds to computing the (minimum-size) prime implicants that
explain the prediction as a subset of the point in feature space.
This relationship is detailed next.

\paragraph{General Case.}
In a more general setting, features can take arbitrary real,
integer or Boolean values. In this case we consider an input cube $C$
and a prediction $\fml{E}$.
The relationship between abductive explanations and prime implicants
is well-known (e.g.~\cite{marquis-fair91,marquis-hdrums00}.
Let $C$ be the cube associated with some satisfying assignment.
Regarding the computation of abductive explanations,
$C\land\fml{F}\nentails\bot$ and the same holds for any
subset of $C$. This means that we just need to consider the constraint
$C\land\fml{F}\entails\fml{E}$, which is equivalent to
$C\entails(\fml{F}\to\fml{E})$. Thus, a \emph{subset-minimal
  explanation}
$C_m$ (given $C$) is a prime implicant of $\fml{F}\to\fml{E}$ (given
$C$), and a \emph{cardinality-minimal explanation} $C_M$ (given $C$)
is a cardinality-minimal prime implicant of $\fml{F}\to\fml{E}$ (given
$C$).
Thus, we can compute subset-minimal (resp.~cardinality-minimal)
explanations by computing instead prime implicants (resp.~shortest
PIs) of $\fml{F}\to\fml{E}$.
As a final remark, the cardinality minimal prime implicants of
$\fml{F}\to\fml{E}$ are selected among those that are contained in
$C$.

\paragraph{Computing Explanations.}
This section  outlines the algorithms for computing a subset-minimal
explanation and a cardinality-minimal explanation.
The computation of a subset-minimal explanation requires a linear
number of calls to a $\fml{T}$-oracle. In contrast, the computation of
a cardinality-minimal explanation in the general case requires a
worst-case exponential number of calls to a
$\fml{T}$-oracle~\cite{liberatore-aij05}.

\begin{algorithm}[!t]
  \caption{Computing a subset-minimal explanation} \label{alg:smexpl}

  \DontPrintSemicolon
  \SetAlgoNoLine
  \LinesNumbered
  \SetFillComment
  \SetKw{KwNot}{not\xspace}
  \SetKw{KwAnd}{and\xspace}
  \SetKw{KwOr}{or\xspace}
  \SetKw{KwBreak}{break\xspace}
  \SetKwData{false}{{\small false}}
  \SetKwData{true}{{\small true}}
  \SetKwData{st}{{\slshape st}}
  \SetKwData{cores}{$\mathcal{C}$}
  \SetKwFunction{ent}{Entails}
  \SetKwFunction{ip}{IP}
  \SetKwBlock{Let}{let}{end}
  \SetKwBlock{FBlock}{}{end}

  \KwIn{\,\,\,\,$\fml{F}$ under $\fml{M}$, initial cube $C$, prediction $\fml{E}$}
  \KwOut{Subset-minimal explanation $C_m$}
  \BlankLine
  \Begin{
    \SetAlgoVlined
    \BlankLine
    \ForEach{$l \in C$}{
      \If{$\ent(C\setminus\{l\},\fml{F}\rightarrow\fml{E},\fml{M})$}{
        $C\gets C\setminus\{l\}$
      }
    }
    \BlankLine
    \Return $C$ \;
  }
  \BlankLine
\end{algorithm}

Algorithm~\ref{alg:smexpl} shows the algorithm to compute a subset-minimal
explanation for a prediction $\fml{E}$ made by an ML model $\mathbb{M}$ encoded
into formula $\fml{F}$ under model $\fml{M}$.
Given a cube $C$ encoding a data sample for the prediction $\fml{E}$, the
procedure returns its minimal subset $C_m$ s.t.\ $\fml{F}\land
C_m\entails\fml{E}$ under $\fml{M}$.
Based on the observation made above, Algorithm~\ref{alg:smexpl} iteratively tries to
remove literals $l$ of the input cube $C$ followed by a check whether the
remaining subcube is an implicant of formula $\fml{F}\rightarrow\fml{E}$, i.e.\
$(C\setminus\{l\})\entails(\fml{F}\rightarrow\fml{E})$.
Note that to check the entailment, it suffices to test whether formula
$(C\setminus\{l\})\land\fml{F}\land\neg{\fml{E}}$ is \emph{false}.
As a result, the algorithm traverses all literals of the cube and, thus, makes
$|C|$ calls to the $\fml{T}$-oracle.

\begin{algorithm}[!t]
  \caption{Computing a smallest size explanation} \label{alg:cmexpl}

  \DontPrintSemicolon
  \SetAlgoNoLine
  \LinesNumbered
  \SetFillComment
  \SetKw{KwNot}{not\xspace}
  \SetKw{KwAnd}{and\xspace}
  \SetKw{KwOr}{or\xspace}
  \SetKw{KwBreak}{break\xspace}
  \SetKwData{false}{{\small false}}
  \SetKwData{true}{{\small true}}
  \SetKwData{st}{{\slshape st}}
  \SetKwData{cores}{$\mathcal{C}$}
  \SetKwFunction{ent}{Entails}
  \SetKwFunction{ip}{IP}
  \SetKwBlock{Let}{let}{end}
  \SetKwBlock{FBlock}{}{end}

\SetKwFunction{SAT}{SAT}
\SetKwFunction{CNF}{CNF}
\SetKwFunction{minhs}{MinimumHS}
\SetKwFunction{model}{GetAssignment}
\SetKwFunction{falselits}{PickFalseLits}

\KwIn{\,\,\,\,$\fml{F}$ under $\fml{M}$, initial cube $C$, prediction $\fml{E}$}
  \KwOut{Cardinality-minimal explanation $C_M$}
  \BlankLine
  \Begin{
    \SetAlgoVlined
    $\Gamma\gets \emptyset$ \label{ln:init}\;
    \While{\true}{
      $h\gets\minhs(\Gamma)$ \label{ln:maxmod} \label{ln:hset}\;
      \If{$\ent(h,\fml{F}\rightarrow\fml{E},\fml{M})$\label{ln:check}}{\Return $h$}
      \Else{
        $\mu\gets\model()$ \label{ln:model}\;
        $C'\gets\falselits(C\setminus h, \mu)$ \label{ln:falsified}\;
        $\Gamma\gets \Gamma\cup C'$ \label{ln:update}\;
      }
    }
  }
  \BlankLine
\end{algorithm}

Computing a cardinality-minimal explanation is hard for
$\stwop$~\cite{eiter-jacm95} and so it is practically less efficient to
perform.
For the propositional case, a smallest size explanation can be extracted using
directly a propositional abduction solver~\cite{imms-ecai16} applied to the
setup described above.
Note that in the general case dealing with FOL formulas, the number of all
possible hypotheses is infinite and so a similar setup is not applicable.
However, and analogously to Algorithm~\ref{alg:smexpl}, one can start with a given
cube $C$ representing an input data sample and consistent with formula
$\fml{F}$ representing an ML model.

Algorithm~\ref{alg:cmexpl} shows a pseudo-code of the procedure computing a smallest
size explanation for prediction $\fml{E}$.
The algorithm can be seen as an adaptation of the propositional abduction
approach~\cite{imms-ecai16} to the general ML explanation problem and is
based on the implicit hitting set paradigm.
As such, Algorithm~\ref{alg:cmexpl} is an iterative process, which deals with a set
$\Gamma$ of the sets to hit.
Initially, set $\Gamma$ is empty (line~\ref{ln:init}).
At each iteration, a new smallest size hitting set $h$ for $\Gamma$ is computed
(see line~\ref{ln:hset}).
Hitting set $h$ is then treated as a cube, which is tested on whether it is a
prime implicant of formula $\fml{F}\rightarrow\fml{E}$ under model $\fml{M}$.
As was discussed above, this can be tested by calling a $\fml{T}$-oracle on
formula $h\land\fml{F}\land\neg{\fml{E}}$ (line~\ref{ln:check}).
If the $\fml{T}$-oracle returns \emph{false}, the algorithm reports hitting set
$h$ as a smallest size explanation for the prediction and stops.
Otherwise, i.e.\ if the $\fml{T}$-oracle returns \emph{true}, an assignment
$\mu$ for the free variables of formula $h\land\fml{F}\land\neg{\fml{E}}$ is
extracted (see line~\ref{ln:model}).
Assignment $\mu$ is then used to determine a subset $C'$ of literals of cube
$C\setminus h$ that were falsified by the previous call to the
$\fml{T}$-oracle.
Finally, set $\Gamma$ is updated on line~\ref{ln:update} to include $C'$ and the
process continues.

Note that the correctness of Algorithm~\ref{alg:cmexpl}, although not proved here,
immediately follows from the correctness of the original hitting set based
approach to propositional abduction~\cite{imms-ecai16}.
The intuition behind the algorithm is the following.
Every iteration checks whether a given (smallest size) subset $h$ of the input
cube $C$ is an implicant of $\fml{F}\rightarrow\fml{E}$.
If this is not the case, some other literals, i.e.\ from set $C\setminus h$,
should be included to $h$ at the next iteration of the algorithm.
Moreover, a new set $C'$ to hit comprises only literals of $C\setminus h$ that
were falsified during the previous $\fml{T}$-oracle call because they are
guaranteed to have been disabled previously, not only by our choice of $h$
but also by the $\fml{T}$-oracle call.



\begin{example}
Consider an example model mapping pairs of all possible integers $i_1$ and
$i_2$ into set $\{1,2\}$.
Assume the model is encoded to the following set of (indicator) constraints
$\fml{F}$ given variables $z_k\in\{0,1\}$, $k\in[3]$:
\begin{center}
  $z_1 = 1 \leftrightarrow i_1 \leq 0,\,\, z_3 = 1 \leftrightarrow z_1+z_2 \leq 1,\,\, z_3 =
  1 \rightarrow o = 1$ \\
  $z_2 = 1 \leftrightarrow i_2 \leq 0,\;\;\;\;\;\;\;\;\;\;\;\;\;\;\;\;\;\;\;\;\;\;\;\;\;\;\;\;\;\;\;\;\;\;\;\, z_3 = 0 \rightarrow o = 2$ \\
\end{center}
Given a data sample encoded as a cube $C=(i_1=3)\land(i_2=2)$, the prediction
is $1$.
It is clear that there are two minimal explanations for this classification:
$C_m^1=(i_1=3)$ and $C_m^2=(i_2=2)$.
Observe that both can be trivially computed by the linear search procedure of
Algorithm~\ref{alg:smexpl} because $C_m^j\entails(\fml{F}\rightarrow(o=1))$ is
\emph{true} for $j\in[2]$.
\end{example}

\newcommand{\NN}{\mbox{\sc NN}}
\newcommand{\relu}{\mbox{\sc ReLU}}
\def \R {\mathbb{R}}

\section{Encoding Neural Networks with MILP} 
This section considers a MILP encoding of a neural network with a commonly-used `rectified linear unit'  nonlinear operator ($\relu$). For simplicity, we explain an encoding of a building block of the network as all blocks have the same structure and are assembled sequentially to form the network. A block consists of a linear transformation and a non-linear transformation.
Let $x = \{x_1,\ldots,x_n\}, x \in \R^n $ be an input and $y=\{y_1, \ldots, y_m\}\in \R_{\geq 0}^m$ be an output of a block.
 First, we apply a linear transformation $x' = Ax + b$, where $A \in \R^m\times \R^n$ and $b \in \R^m$
are real-valued parameters of the network. Then we apply a non-linear transformation $y = \relu(x')$, where $\relu = \mbox{max}(x',0)$.

To encode a block, we  use a recently proposed MILP encoding~\cite{fischetti-cj18}. One useful modeling property of this encoding is that it employs indicator constraints that are natively supported by modern MILP solvers. Hence, we can avoid using the Big-M notation in the encoding that requires good bounds tightening to compute the value of Big-M.
To perform the encoding, we introduce two sets of variables: Boolean variables $z = \{z_1, \ldots, z_m\}, z \in \{0,1\}^m$ and auxiliary real variables $s=\{s_1, \ldots, s_m\}, s \in \R_{\geq 0}^m$. Intuitively,
the $z_i$ variable encodes the sign of  $\sum_{j=1}^na_{i,j}x_j + b_i$. If $z_i = 1$ then $\sum_{j=1}^na_{i,j}x_j + b_i \leq 0$ and $y_i = 0$. If $z_i = 0$ then $\sum_{j=1}^na_{i,j}x_j  + b_i  \geq 0$ and $y_i = \sum_{j=1}^na_{i,j}x_j  + b_i $. A block is encoded as follows:
\begin{align}
\sum_{j=1}^na_{i,j}x_j  + b_i= y_i - s_i, \label{eq:enc_lin} \\
z_i = 1  \rightarrow  y_i \leq 0,  \label{eq:sign_pos}\\
z_i = 0  \rightarrow  s_i \leq 0,  \label{eq:sign_new}\\
y_i \geq 0,  s_i \geq 0, z_i \in \{0,1\},  \label{eq:domains}
\end{align}
\noindent where $i \in [1,m]$. To see why the encoding is correct we consider two cases. First, we consider the case $z_i = 1$. As we mentioned above, $y_i$ must be zero in this case. Indeed, if $z=1$ then $ y_i \leq 0$ holds. Together with
$y_i \geq 0$, we get that $y_i =0$. Similarly,  $z_i = 0$ should ensure that $\sum_{j=1}^na_{i,j}x_j  + b_i  = y_i$.
If $z=0$ then $ s_i \leq 0$ forcing $s_i = 0$. In this case, $y_i$ equals to $\sum_{j=1}^na_{i,j}x_j + b_i $.

A common neural network  architecture for classification problems performs a normalization of the network output as the last layer, e.g. the softmax layer. However, we do not need to encode this normalization transformation as it does not change the maximum value of the output that defines prediction of the network.

\begin{example}
Consider an example of  a block with two inputs, $x_1$ and $x_2$ and two outputs $y_1$ and $y_2$. Let $A = [2, -1; 1, 1]$ and $b = [-1, 1]$ be parameters of a linear transformation. To encode this block, we introduce auxiliary variables $s_1$, $s_2$, $z_1$ and $z_2$. We obtain the following constraints:
\begin{align*}
2x_1  - x_2 - 1= y_1 - s_1, \\
x_1 +  x_2  + 1= y_2 - s_2, \\
z_1 = 1  \rightarrow  y_1 \leq 0, z_2 = 1  \rightarrow  y_2 \leq 0, \\
z_1 = 0  \rightarrow  s_1 \leq 0, z_2 = 0  \rightarrow  s_2 \leq 0,  \\
y_1 \geq 0, y_2 \geq 0,  s_1 \geq 0, s_2 \geq 0, z_1 \in \{0,1\}, z_2 \in \{0,1\}.
\end{align*}
\end{example}
%



\section{Experimental Results} \label{sec:res} 
%
%
This section evaluates the scalability of the proposed approach
to computing cardinality- and subset-minimal explanations and the quality of
the computed explanations (in terms of the minimality guarantees).
The benchmarks considered include the well-known text-based datasets from the
UCI Machine Learning Repository\footnote{\url{https://archive.ics.uci.edu/ml/}}
and Penn Machine Learning
Benchmarks\footnote{\url{https://github.com/EpistasisLab/penn-ml-benchmarks/}},
as well as the widely used MNIST digits
database\footnote{\url{http://yann.lecun.com/exdb/mnist/}}.

\paragraph{Setup and prototype implementation.} \label{sec:tool}
To assess scalability, all benchmarks were ran on a Macbook Pro having
an Intel Core i7\,2.8GHz processor with 8GByte of memory on board.
Time limit was set to 1800 seconds while memory limit was set to
4GByte.
The prototype implementation of the proposed approach follows
Algorithm~\ref{alg:smexpl} and Algorithm~\ref{alg:cmexpl} for computing subset-
and cardinality-minimal explanations, respectively.
It is written in Python and targets both SMT and MILP
solvers\footnote{Here we mean that training and encoding of neural
networks as well as the explanation procedures are implemented in
Python.}.
SMT solvers are accessed through the PySMT framework~\cite{gm-smt15},
which
%
provides a unified interface to SMT solvers like CVC4, MathSAT5,
Yices2, and Z3 among a few others.
Note that in the following only the results of
Yices2~\cite{dutertre-cav14} are shown as of the best performing SMT
solver, selected based on a prior experimentation with CVC4, MathSAT5,
Yices2, and Z3.
CPLEX\,12.8.0~\cite{cplex-page} is used as a MILP oracle accessed via
its official Python API.\footnote{Note that more efficient reasoning
  engines
  exist~\cite{barrett-cav17,kwiatkowska-ijcai18,vechev-sp18,tiwari-nfm18}.
  Those were not tested because the explanation procedure relies on
  efficient incremental access to the oracle and its ability to add
  and remove constraints \emph{``on the fly''}.}
The implementation of minimum hitting set enumeration in
Algorithm~\ref{alg:cmexpl} is based on an award-winning maximum
satisfiability solver
RC2\footnote{\url{https://maxsat-evaluations.github.io/2018}} written
on top of the PySAT toolkit~\cite{imms-sat18}.

\paragraph{Quality of explanations.} \label{sec:quality}
This section focuses on a selection of datasets from UCI Machine Learning
Repository and Penn Machine Learning Benchmarks (see Table~\ref{tab:penn-ml} for
details).
%
%
The selected datasets have 9--32 features and contain 164--691 data samples.
The experiment is organized as follows.
First, given a dataset, a neural network is trained\footnote{Each neural
network considered has one hidden layer  with $i\in\{10,15,20\}$ neurons. The
accuracy of the trained NN classifiers is at least $92\%$ on all the
considered datasets.} and encoded as described above.
Second, the explanation procedure (computing either a subset- or a
cardinality-minimal explanation) is ran for each sample of the dataset.
If all samples get explained within 1800 seconds in total, the procedure is
deemed to succeed.
Otherwise, the process is interrupted meaning that the explanations for some of
the samples are not extracted successfully.

Column~1 of Table~\ref{tab:penn-ml} lists the selected datasets followed by the
number of features. Column~4 details the minimal, average, and maximal size of
explanations per sample for a given a dataset (depending on prefix \textbf{m},
\textbf{a}, and \textbf{M} in column~3).
Analogously, columns 5--9 depict the minimal, average, and maximal time spent
for computing an explanation for a given dataset, either with an SMT or a MILP
oracle in use.

As one can observe, using a MILP oracle is preferred as it consistently
outperforms its SMT rival.
In general, the MILP-based solution is able to compute a subset-minimal
explanation of a sample in a fraction of a second.
Also note that subset- and cardinality-minimal explanation size varies a lot
depending on the dataset.
On the one hand, it may be (see \emph{australian}) enough to keep just 1
feature to explain the outcome, which is $\approx 7\%$ of the data sample.
On the other hand, some data samples cannot be reduced at all (see the
\textbf{M} values in column~4).
On average, the relative size of subset-minimal explanations varies from
$28.5\%$ to $86.7\%$ with the mean value being $60.5\%$.
This is deemed to provide a reasonable reduction of sample size, which may
help a human interpret the outcomes of a machine learning classifier.

\begin{table}[!t]
  \caption{Subset-minimal and cardinality-minimal explanations for the selected
    UCI and PennML datasets. Explanations are computed with the use of an SMT
    or a MILP oracle. The data shows minimal (\textbf{m}), average
    (\textbf{a}), and maximal (\textbf{M}) size of the computed explanations
    per dataset, as well as minimal (\textbf{m}), average (\textbf{a}), and
  maximal (\textbf{M}) time per dataset required to extract one explanation
either by an SMT oracle or its MILP counterpart.}
  \label{tab:penn-ml}
  \scriptsize
  \begin{adjustbox}{center}
    \begin{tabular}{c@{\hspace{4pt}}c@{\hspace{4pt}}c@{\hspace{4pt}}c@{\hspace{4pt}}c@{\hspace{4pt}}c@{\hspace{4pt}}c@{\hspace{4pt}}c@{\hspace{4pt}}c@{\hspace{4pt}}c@{\hspace{4pt}}}
      \toprule
      \multirow{2}{*}{\textbf{Dataset}} &
                                                      & &
      \multicolumn{3}{c}{{\textbf{Minimal explanation}}} &
      \multicolumn{3}{c}{{\textbf{Minimum explanation}}} \\
      \cmidrule(lr){4-9}
      & & & \textbf{size} & \textbf{SMT (s)} & \textbf{MILP (s)} & \textbf{size} & \textbf{SMT (s)} & \textbf{MILP (s)} \\
      \midrule
      \multirow{3}{*}{australian} &
      \multirow{3}{*}{\scriptsize($14$)} &
      \textbf{m} &
        $1$ &
        $0.03$ &
        $0.05$ &
        --- &
        --- &
        --- \\
      & & \textbf{a} &
        $8.79$ &
        $1.38$ &
        $0.33$ &
        --- &
        --- &
        --- \\
      & & \textbf{M} &
        $14$ &
        $17.00$ &
        $1.43$ &
        --- &
        --- &
        --- \\
      \midrule
      \multirow{3}{*}{auto} &
      \multirow{3}{*}{\scriptsize($25$)} &
      \textbf{m} &
        $14$ &
        $0.05$ &
        $0.19$ &
        --- &
        --- &
        --- \\
      & & \textbf{a} &
        $19.99$ &
        $2.69$ &
        $0.36$ &
        --- &
        --- &
        --- \\
      & & \textbf{M} &
        $25$ &
        $37.38$ &
        $0.76$ &
        --- &
        --- &
        --- \\
      \midrule
      \multirow{3}{*}{backache} &
      \multirow{3}{*}{\scriptsize($32$)} &
      \textbf{m} &
        $13$ &
        $0.13$ &
        $0.14$ &
        --- &
        --- &
        --- \\
      & & \textbf{a} &
        $19.28$ &
        $5.08$ &
        $0.85$ &
        --- &
        --- &
        --- \\
      & & \textbf{M} &
        $26$ &
        $22.21$ &
        $2.75$ &
        --- &
        --- &
        --- \\
      \midrule
      \multirow{3}{*}{breast-cancer} &
      \multirow{3}{*}{\scriptsize($9$)} &
      \textbf{m} &
        $3$ &
        $0.02$ &
        $0.04$ &
        $3$ &
        $0.02$ &
        $0.03$ \\
      & & \textbf{a} &
        $5.15$ &
        $0.65$ &
        $0.20$ &
        $4.86$ &
        $2.18$ &
        $0.41$ \\
      & & \textbf{M} &
        $9$ &
        $6.11$ &
        $0.41$ &
        $9$ &
        $24.80$ &
        $1.81$ \\
      \midrule
      \multirow{3}{*}{cleve} &
      \multirow{3}{*}{\scriptsize($13$)} &
      \textbf{m} &
        $4$ &
        $0.05$ &
        $0.07$ &
        $4$ &
        --- &
        $0.07$ \\
      & & \textbf{a} &
        $8.62$ &
        $3.32$ &
        $0.32$ &
        $7.89$ &
        --- &
        $5.14$ \\
      & & \textbf{M} &
        $13$ &
        $60.74$ &
        $0.60$ &
        $13$ &
        --- &
        $39.06$ \\
      \midrule
      \multirow{3}{*}{cleveland} &
      \multirow{3}{*}{\scriptsize($13$)} &
      \textbf{m} &
        $7$ &
        --- &
        $0.07$ &
        $7$ &
        --- &
        $0.07$ \\
      & & \textbf{a} &
        $9.23$ &
        --- &
        $0.34$ &
        $8.92$ &
        --- &
        $2.03$ \\
      & & \textbf{M} &
        $13$ &
        --- &
        $1.02$ &
        $13$ &
        --- &
        $12.25$ \\
      \midrule
      \multirow{3}{*}{glass} &
      \multirow{3}{*}{\scriptsize($9$)} &
      \textbf{m} &
        $4$ &
        $0.02$ &
        $0.05$ &
        $4$ &
        $0.02$ &
        $0.05$ \\
      & & \textbf{a} &
        $7.82$ &
        $0.18$ &
        $0.15$ &
        $7.68$ &
        $1.13$ &
        $0.29$ \\
      & & \textbf{M} &
        $9$ &
        $3.26$ &
        $0.38$ &
        $9$ &
        $15.47$ &
        $2.48$ \\
      \midrule
      \multirow{3}{*}{glass2} &
      \multirow{3}{*}{\scriptsize($9$)} &
      \textbf{m} &
        $2$ &
        $0.02$ &
        $0.05$ &
        $2$ &
        $0.02$ &
        $0.04$ \\
      & & \textbf{a} &
        $5.20$ &
        $0.47$ &
        $0.15$ &
        $4.51$ &
        $3.31$ &
        $0.74$ \\
      & & \textbf{M} &
        $9$ &
        $1.42$ &
        $0.38$ &
        $9$ &
        $10.98$ &
        $2.21$ \\
      \midrule
      \multirow{3}{*}{heart-statlog} &
      \multirow{3}{*}{\scriptsize($13$)} &
      \textbf{m} &
        $3$ &
        $0.01$ &
        $0.05$ &
        $3$ &
        $0.01$ &
        $0.03$ \\
      & & \textbf{a} &
        $8.13$ &
        $0.13$ &
        $0.15$ &
        $7.09$ &
        $1.51$ &
        $1.01$ \\
      & & \textbf{M} &
        $13$ &
        $0.89$ &
        $0.38$ &
        $13$ &
        $8.16$ &
        $8.12$ \\
      \midrule
      \multirow{3}{*}{hepatitis} &
      \multirow{3}{*}{\scriptsize($19$)} &
      \textbf{m} &
        $6$ &
        $0.02$ &
        $0.04$ &
        $4$ &
        $0.01$ &
        $0.04$ \\
      & & \textbf{a} &
        $11.42$ &
        $0.07$ &
        $0.06$ &
        $9.39$ &
        $4.07$ &
        $2.89$ \\
      & & \textbf{M} &
        $19$ &
        $0.26$ &
        $0.20$ &
        $19$ &
        $27.05$ &
        $22.23$ \\
      \midrule
      \multirow{3}{*}{voting} &
      \multirow{3}{*}{\scriptsize($16$)} &
      \textbf{m} &
        $3$ &
        $0.01$ &
        $0.02$ &
        $3$ &
        $0.01$ &
        $0.02$ \\
      & & \textbf{a} &
        $4.56$ &
        $0.04$ &
        $0.13$ &
        $3.46$ &
        $0.3$ &
        $0.25$ \\
      & & \textbf{M} &
        $11$ &
        $0.10$ &
        $0.37$ &
        $11$ &
        $1.25$ &
        $1.77$ \\
      \midrule
      \multirow{3}{*}{spect} &
      \multirow{3}{*}{\scriptsize($22$)} &
      \textbf{m} &
        $3$ &
        $0.02$ &
        $0.02$ &
        $3$ &
        $0.02$ &
        $0.04$ \\
      & & \textbf{a} &
        $7.31$ &
        $0.13$ &
        $0.07$ &
        $6.44$ &
        $1.61$ &
        $0.67$ \\
      & & \textbf{M} &
        $20$ &
        $0.88$ &
        $0.29$ &
        $20$ &
        $8.97$ &
        $10.73$ \\
      \bottomrule
    \end{tabular}
  \end{adjustbox}
\end{table}

\paragraph{Subset- vs.\ cardinality-minimal explanations.} \label{sec:svsc}
Compared to subset-minimal explanations, computing smallest size explanations
is significantly more expensive due to the problem being hard for the second
level of the polynomial hierarchy.
As one can observe, the proposed explanation procedure fails to explain
\emph{all} data samples within the given total 1800 seconds (see
\emph{australian}, \emph{auto}, \emph{backache}).
As in the case of minimal explanations, the MILP oracle outperforms the
SMT-based solver being able to explain 2 more datasets.
The size of smallest size explanations varies from $21.6\%$ to $85.3\%$ with
the average value being $52.6\%$.
Although cardinality-minimal explanations are in general smaller than
subset-minimal ones, their computation takes a lot more time and so the overall
advantage of minimum size explanations seems questionable.

\paragraph{State-of-the-art in logic-based explanations.} \label{sec:votes}
As a side problem, here we compare the quality of explanations produced by the
current approach with the state of the art in logic-based explanation of
Bayesian network classifiers~\cite{darwiche-ijcai18}, given a concrete dataset.
Following~\cite{darwiche-ijcai18}, let us focus on the Congressional Voting
Records dataset (referred to as \emph{voting} in Table~\ref{tab:penn-ml}).
The dataset contains 16 key votes by Congressmen of the U.S.\ House of
Representatives, expressed as Boolean \emph{yes} and \emph{no} (\emph{1} and
\emph{0}).
Consider the following list of votes classified as Republican:
\begin{center}
  \texttt{(0~1~0~1~1~1~0~0~0~0~0~0~1~1~0~1)}
\end{center}

The BDD-based approach of~\cite{darwiche-ijcai18} for explaining Bayesian
network classifiers computes the following \emph{smallest size} explanations of
size~9:
\begin{center}
  \texttt{(~~~~0~1~1~~~0~0~0~~~~~~~1~1~0~~)}

  \texttt{(~~~~0~1~1~1~~~0~0~~~~~~~1~1~0~~)}
\end{center}

To be able to compare to this data, we trained 4 neural networks separately of
each other\footnote{Due to randomization when training, the resulting networks
may represent different functions and, thus, behave differently.} and tried to
``explain'' their predictions given this concrete data sample.
As a result, we got the following \emph{subset-minimal} explanations of size
varying from 3 to 5:
\begin{center}
  \texttt{(~~~~~~1~~~~~~~~~0~~~0~~~~~~~0~~)}

  \texttt{(~~~~~~1~~~~~~~~~0~~~0~~~~~~~~~~)}

  \texttt{(~~~~0~1~~~~~~~~~0~~~0~~~~~~~0~~)}

  \texttt{(~~~~0~1~~~~~~~~~0~~~0~~~~~~~~~1)}
\end{center}

A possible intuition behind this impressive result is that models based on
neural networks may generalize better than solutions relying on Bayesian
networks and/or they allow for more ``aggressive'' and, thus, more efficient
interpretation.

It should also be noted that the proposed approach is constraint-agnostic and
computes explanations ``of the fly'' in the \emph{online} manner while the work
of~\cite{darwiche-ijcai18} relies on prior compilation of the classifier
function into a BDD, which is in general known to be computationally expensive.

\paragraph{Scalability and MNIST digits.} \label{sec:mnist}
A widely used benchmark dataset for testing machine learning models is MNIST
digits, which comprises a set of greyscale images depicting hand-written
digits.
This section aims at illustrating visually and discussing a few minimal
explanations provided for example predictions of a neural network trained on
the MNIST digits datasets.
Concretely, let us consider a neural network with one hidden layer containing
15 or 20 neurons trained to distinguish two digits, e.g.\ 1 and 3, or 1 and 7,
among other pairs.
Each MNIST data sample describes a picture of size $\text{28}\times\text{28}$, and so the
total number of features is 784.

Observe that the number of features makes these datasets significantly more
challenging for extracting explanations.
Our experiments confirm this fact as the average time spent for computing one
subset-minimal explanation using a MILP oracle is about $52.86$ seconds.
As for SMT, given 1~hour time limit, we were unable to get an explanation for
any MNIST data sample using the SMT solvers available in PySMT.
The average size of subset-minimal explanations varies from $46.4\%$ to
$80.3\%$ of the total number of pixels (i.e.\ $784$), with the mean value being
$63.92\%$.
Also note that no cardinality-minimal explanation was computed within 1~hour
time limit. This holds for all SMT and MILP alternatives tried.

\begin{figure}[!t]
  \centering
  \begin{subfigure}[b]{0.10\textwidth}
    \centering
    \includegraphics[width=\textwidth]{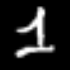}
    \caption{}
    \label{fig:1o}
  \end{subfigure}%
  \;\;
  \begin{subfigure}[b]{0.10\textwidth}
    \centering
    \includegraphics[width=\textwidth]{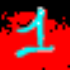}
    \caption{}
    \label{fig:1s}
  \end{subfigure}
  \;
  \begin{subfigure}[b]{0.10\textwidth}
    \centering
    \includegraphics[width=\textwidth]{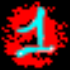}
    \caption{}
    \label{fig:1c}
  \end{subfigure}
  \;
  \begin{subfigure}[b]{0.10\textwidth}
    \centering
    \includegraphics[width=\textwidth]{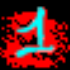}
    \caption{}
    \label{fig:1l}
  \end{subfigure}
  \caption{Possible minimal explanations for digit one.}
  \label{fig:expl-one}
\end{figure}

\begin{figure}[!t]
  \centering
  \begin{subfigure}[b]{0.10\textwidth}
    \centering
    \includegraphics[width=\textwidth]{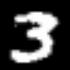}
    \caption{}
    \label{fig:3o}
  \end{subfigure}%
  \;\;
  \begin{subfigure}[b]{0.10\textwidth}
    \centering
    \includegraphics[width=\textwidth]{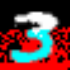}
    \caption{}
    \label{fig:3s}
  \end{subfigure}
  \;
  \begin{subfigure}[b]{0.10\textwidth}
    \centering
    \includegraphics[width=\textwidth]{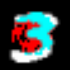}
    \caption{}
    \label{fig:3c}
  \end{subfigure}
  \;
  \begin{subfigure}[b]{0.10\textwidth}
    \centering
    \includegraphics[width=\textwidth]{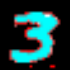}
    \caption{}
    \label{fig:3l}
  \end{subfigure}
  \caption{Possible minimal explanations for digit three.}
  \label{fig:expl-three}
\end{figure}

Let us focus on two particular data samples shown in Figure~\ref{fig:1o} and
Figure~\ref{fig:3o}.
These samples describe concrete ways of writing digits 1 and 3.
Subset-minimal explanations for these samples computed by Algorithm~\ref{alg:smexpl}
are depicted in Figure~\ref{fig:1s} and Figure~\ref{fig:3s}, respectively.
(Observe that pixels included in the explanations are either red or magenta
while greyscale parts of the images are excluded from the explanations.)
Note that Algorithm~\ref{alg:smexpl} tries to remove pixels from an image one by one
starting from the top left corner and ending at the bottom right corner.
As shown in Figure~\ref{fig:1s} and Figure~\ref{fig:3s}, this procedure ends up
keeping the bottom part of the image as it is needed to forbid
misclassifications while the top part of the image being removed.
While this is perfectly correct from the logical perspective, it may not
satisfy some users as it does not give any reasonable hint on why the image is
classified that specific way.

In this situation, better results could be obtained if the pixels were
traversed in a different order.
For instance, one may find reasonable first to remove pixels that are far from
the image center because the most important information in an image is
typically located close to its center.
Applying this strategy to the considered samples results in the explanations
highlighted in Figure~\ref{fig:1c} and Figure~\ref{fig:3c}.

Another possible policy would be to prefer removing dark parts of an image
keeping the light parts.
The result of applying this policy is shown in Figure~\ref{fig:1l} and
Figure~\ref{fig:3l}.\footnote{Note that none of the presented explanations for
  digit~1 follows exactly the shape of the digit. The reason is that in many
  cases the area occupied by the shape of digit~1 is \emph{``contained''} in
  the area of some shapes of digit~3 and so some of the pixels
\emph{``irrelevant''} to~1 are needed in order to forbid classifying the digit
as~3.}
Moreover, a user may consider other heuristics to sort the pixels including
various combinations of the ones described above.
As there can be myriads of possible explanations for the same data sample,
trying various heuristics to sort/order the pixels can help obtaining
explanations that are more interpretable than the other from a human's point of
view.

\paragraph{Summary.} \label{sec:summary}
As shown above, the proposed approach for computing subset- and
cardinality-minimal explanations for ML models using an underlying constraints
formalism can obtain high quality results for a variety of popular datasets,
e.g.\ from UCI Machine Learning Repository, Penn Machine Learning Benchmarks,
and also MNIST digits.
In particular, the experimental results indicate that the approach, if applied
to neural network classifiers, is capable of obtaining explanations that are
reasonably smaller than those of the state of the art and, thus, may be
preferred as more interpretable from the point of view of human decision
makers.

Scalability of the proposed ideas was tested with the use of SMT and MILP
oracles.
As it was shown, MILP in general outperforms its SMT counterparts.
It was also shown that computing smallest-size explanations is computationally
quite expensive, which makes subset-minimal explanations a better alternative.
Another important factor here is that subset-minimal explanations are usually
just a bit larger than cardinality-minimal explanations.




\section{Related Work}

We briefly overview two lines of research on heuristic-based explanations of machine learning models. The first line focuses on explaining a complex ML model as a whole using an interpretable model~\cite{hinton-cexaiia17,ZhangCNN_DT2018}.  Such algorithms take a pair of ML models, an original complex ML model and a target explainable model, as an input. For example, the original model may be a trained neural network, and the target model may be a decision tree~\cite{ZhangCNN_DT2018}.  One of the underlying assumptions here is that there exists a transformation from the original model to the target model that preserves the accuracy of the original model.  The second line of work focuses on explanations of an ML model for a given input~\cite{guestrin-kdd16,guestrin-aaai18,SimonyanVZ13}. For example, given an image of a car, we might want an explanation why a neural network classifies this image as a car. The most prominent example is the LIME framework~\cite{guestrin-kdd16}. The main idea of the method is to learn important features of the input using its local perturbations. The algorithm observes how the ML model behaves on perturbed inputs. Based on this information, important features for the classification of a given sample are linearly separated from the rest of the features. The LIME method is model agnostic, as it can work with any ML model. Another group of methods is based on the saliency map~\cite{SimonyanVZ13}. The idea is that the most important input features can be  obtained using the knowledge of the ML model, e.g. gradients. However, none of these approaches provides any guarantees on the quality of explanations.



\section{Discussion and Future Work} \label{sec:futw}

This paper proposes a principled approach to the problem of computing
explanations of ML models. 
Nevertheless, the use of abductive reasoning, and the associated
complexity, can raise concerns regarding \emph{robustness} and
\emph{scalability}.
%
Several ways can be envisioned to tackle the robustness issue.
First, the approach enables a user not only to compute one
explanation, either minimal or minimum, but also to \emph{enumerate} a
given number of such explanations.
Second, since the proposed approach is based on calls to some oracle,
preferences over explanations, e.g.\ in terms of (in)dependent feature
weights, can in principle be accommodated.

Regarding the issue of scalability, it should be noted that developed
prototype serves as a \emph{proof of concept} that the proposed
generic approach can provide small and reasonable explanations, which
are arguably easier to understand by a human decision maker, this
despite the underlying ML model being a blackbox, whose decisions may
look uninterpretable per se.
As the experimental results show, this holds for all the benchmark
sets studied. 
Moreover and as the experimental results suggest, the approach still
applies (with no conceptual modification) if one opts to use oracles
other than an ILP
solver~\cite{barrett-cav17,kwiatkowska-ijcai18,vechev-sp18,tiwari-nfm18}.
Given the performance results obtained by these recent works, one
should expect to be able to handle larger-scale NNs.
Furthermore, since the proposed approach is constraint-agnostic, the
same ideas can provide the basis for the the development of efficient
decision procedures and effective encodings, not only for NN-based
classifiers, but also for other state-of-the-art machine learning
models.
%
%
An alternative way to improve scalability would be to apply
\emph{abstraction refinement} techniques, a prominent approach that
proved invaluable in many other settings.

Although the scalability of approaches for computing explanations is
an important issue, the existence of a principled approach can serve
to benchmark other heuristic approaches proposed in recent
work~\cite{guestrin-kdd16,hinton-cexaiia17}.
%
More importantly, the proposed principled approach can elicit new
heuristic approaches, that scale better than the exact methods
outlined in the paper, but which in practice perform comparably to
those exact methods.
Observe also that explanations provided by any heuristic approach can,
not only be validated, but can also be \emph{minimized further}, in
case additional minimization can be achieved.

As a final remark, note that by providing a high-level description,
independent of a concrete ML model, the paper shows how the same
general approach can be applied to any ML model that can be
represented with (first-order logic) constraints.
This demonstrates the flexibility of the proposed solution in that
most existing ML models can (at least conceptually) be encoded with 
FOL.
We emphasize that the approach is perfectly general.
If for some reason FOL is inadequate to encode some ML model, one
could consider second-order or even higher order logics.
Of course, the price to pay would be the (potential) issues with
decidability and scalability.



\section{Conclusions} 

Explanations of ML predictions are essential in a number of
applications.
Past work on computing explanations for ML models has mostly addressed
heuristic approaches, which can yield poor quality solutions.
This paper proposes the use of abductive reasoning, namely the
computation of (shortest) prime implicants, for finding subset- or
cardinality-minimal explanations.
Although the paper considers MILP encodings of NNs, the proposed
approach is constraint-agnostic and independent of the ML model used.
A similar approach can be applied to any other ML model, provided the
model can be represented with some constraint reasoning system and
entailment queries can be decided with a dedicated oracle.
The experimental results demonstrate the quality of the computed explanation,
and highlight an important tradeoff between subset-minimal explanations, which
are computationally easier to find, and cardinality-minimal explanations, which
can be far harder to find, but which offer the best possible quality (measure
as the number of specified features).

A number of lines of work can be envisioned. These include considering
other ML models, other encodings of ML models, other approaches for
answering entailment queries, and also the evaluation of additional
benchmark suites.
%


% ...

\bibliographystyle{aaai}
\bibliography{refs,xrefs}

\end{document}